\documentclass{article}

\usepackage{arxiv}
\usepackage[utf8]{inputenc} 
\usepackage[T1]{fontenc}    
\usepackage{hyperref}       
\usepackage{url}            
\usepackage{booktabs}       
\usepackage{amsfonts}       
\usepackage{nicefrac}       
\usepackage{microtype}      
\usepackage{lipsum}
\usepackage{graphicx}
\usepackage[utf8]{inputenc}
\usepackage[arabic, farsi, english]{babel}
\usepackage{amsmath}
\usepackage{caption}
\usepackage{subcaption}

\usepackage[utf8]{inputenc}
\usepackage{natbib}
\setcitestyle{authoryear, open={(}, close={)}} 

\title{Offensive Language Detection with BERT-based models, By Customizing Attention Probabilities}

\author{
  Peyman Alavi \\
  NLP Research Laboratory \\
  Shahid Beheshti University\\
  Tehran, Iran \\
  \texttt{seyedp.alavi@mail.sbu.ac.ir} \\
  
  \And
  Pouria Nikvand \\
  NLP Research Laboratory \\
  Shahid Beheshti University\\
  Tehran, Iran \\
  \texttt{po.nikvand@mail.sbu.ac.ir} \\
  
  \And
  Mehrnoush Shamsfard \\
  NLP Research Laboratory \\
  Shahid Beheshti University\\
  Tehran, Iran \\
  \texttt{m-shams@sbu.ac.ir} \\
}

\begin{document}
\maketitle

\begin{abstract}
This paper describes a novel study on using `Attention Mask' input in transformers and using this approach for detecting offensive content in both English and Persian languages. The paper's principal focus is to suggest a methodology to enhance the performance of the BERT-based models on the `Offensive Language Detection' task. Therefore, we customize attention probabilities by changing the `Attention Mask' input to create more efficacious word embeddings. To do this, we firstly tokenize the training set of the exploited datasets (by BERT tokenizer). Then, we apply Multinomial Naive Bayes to map these tokens to two probabilities. These probabilities indicate the likelihood of making a text non-offensive or offensive, provided that it contains that token. Afterwards, we use these probabilities to define a new term, namely Offensive Score. Next, we create two separate (because of the differences in the types of the employed datasets) equations based on Offensive Scores for each language to re-distribute the `Attention Mask' input for paying more attention to more offensive phrases. Eventually, we put the F1-macro score as our evaluation metric and fine-tune several combinations of BERT with ANNs, CNNs and RNNs to examine the effect of using this methodology on various combinations. The results indicate that all models will enhance with this methodology. The most improvement was 2\% and 10\% for English and Persian languages, respectively.
\end{abstract}

\keywords{Offensive Language Detection \and BERT-based Models \and Attention Mask \and Attention Probability \and Multinomial Naive Bayes}

\section{Introduction}
These days, social media provide the most convenient way of communication for individuals and make the users capable of sending messages containing attitudes, criticisms and daily conversations immediately. Unfortunately, increasing social media popularity has led to pervading more offensive content to users, which has become a critical problem for the communities \citep{brums}. Insulting content can jeopardise the community’s mental health and can affect user experience detrimentally. Accordingly, it is indispensable to recognise and throttle the offensive content before it appears to individuals’ observance.

The amount of textual information generated daily in social media is tremendous; consequently, it becomes inconceivable and infeasible for humans to identify and remove offensive messages manually. Furthermore, base word filtering could not be an adequate solution because it cannot consider the influence of aspects, such as the domain of an utterance, discourse context, the author's and targeted recipient's identity and other extra information \citep{schmidt-wiegand-2017-survey}. Accordingly, scholars, online communities, social media platforms, and IT companies have decided to elaborate more intelligent systems based on NLP techniques for identifying offensive language.

Along with these efforts, several shared tasks were launched regarding covering a particular aspect of offensive language. TRAC (Trolling, Aggression and Cyberbullying) \citep{trac1-report} \citep{trac2-report}, HatEval \citep{hateval_report_dataset}, HASOC (Hate Speech and Offensive Content Identification) \citep{inproceedingsHASOC}, and OffensEval \citep{offenseval-2019-report} \citep{offenseval-2020-report} are some significant instances of these types of competitions. The first version of OffensEval covered just the English language, while its second version comprised more languages such as Arabic, Danish, Greek and Turkish.

Since the NLP techniques need the numerical representation of textual information, word embedding is widely used. Word embedding is the dense vector representation of words in lower-dimensional space. Several word embedding models have been proposed, which \citeauthor{first_transformer} \citeyearpar{first_transformer} presented a novel neural network architecture, called a transformer, which had many benefits over the conventional sequential models (LSTM, GRU). Bidirectional Encoder Representations from Transformer (BERT) \citep{bert} and ParsBERT \citep{farahani2020parsbert} are such transformers that are used widely for English and Persian word embedding.  

This paper intends to present that the performance of the BERT-based models on the `Offensive Language Detection' task will enhance, provided that the models pay more attention to more offensive phrases. This procedure can be achieved by changing the `Attention Mask' input, which affects the attention probabilities directly. To demonstrate that this improvement occurs in different languages, we tried this methodology on two distinct languages (English and Persian). In English, we used the OLID \citep{olid_dataset} dataset, and in Persian, we created our custom dataset, namely POLID. These two datasets are a little different in context; hence, the approach of changing the `Attention Mask' input will be changed. For each language, we tried several BERT-based models. All these models experienced an improvement in their performance when they paid more attention to more offensive phrases.

The remainder of this paper is structured as follows: section 2 describes the related research and datasets in the field of offensive language detection. Section 3 explicates the core of our methodology. Section 4 presents an analysis of our evaluation results on the OLID \citep{olid_dataset} and POLID datasets. Eventually, Section 5 offers the conclusion. The code of this paper is available here: \url{https://github.com/peyman-alv/offensive-detection}

\section{Background}

\subsection{Related Work}

Fundamentally,  offensive language consists of several varieties, such as aggressive behaviour, bullying and hate speech. \citeauthor{hate_speech_def} \citeyearpar{hate_speech_def} defines hate speech as ``any animosity or disparagement of an individual or a group on account of a group characteristics such as race, colour, national origin, sex, disability, religion, or sexual orientation''. \citeauthor{nitta-etal-2013-detecting} \citeyearpar{nitta-etal-2013-detecting} defines a particular part of bullying, called cyberbullying, as ``humiliating and slandering behaviour towards other people through Internet services, such as BBS, Twitter or e-mails.''

Various text classification approaches have exploited traditional or feature-based supervised learning techniques in early research for covering a particular part of offensive languages \citep{brums}. \citeauthor{6406271} \citeyearpar{6406271} applied text-mining methods to create the Lexical Syntactic Feature (LSF) to identify offensive language in social media and predict a user’s potentiality to send out offensive contents. \citeauthor{malmasi-zampieri-2017-detecting} \citeyearpar{malmasi-zampieri-2017-detecting} applied linear Support Vector Machine (SVM) on several surface n-grams and word skip-grams for detecting hate speech in social media. \citeauthor{razavi} \citeyearpar{razavi} exploited Complement Naïve Bayes, Multinomial Updatable Naïve Bayes and Decision Table/Naïve Bayes hybrid classifier for building a multi-level classifier for flame detection by boosting an underlying dictionary of abusive and insulting phrases. One of the main privileges of their model is being able to be modified based on any accumulative training data. \citeauthor{Montani2018GermEval2} \citeyearpar{Montani2018GermEval2} defined five disjoint sets of features (namely Character N-Grams, Important N-Grams, Token N-Grams, Important Tokens and Word Embeddings) and trained Logistic Regression with balanced class weight, two sets of an ensemble of Random Forests on each feature group and ensemble the results due to detecting abusive language in German. They understood that `Important Tokens' is the most useful feature set because their evaluation metric dropped when this set was removed. Their proposed approach stood at first in GermEval 2018 \citep{germeval}. \citeauthor{indurthi-etal-2019-fermi} \citeyearpar{indurthi-etal-2019-fermi} fed the output of Universal Sentence Encoder \citep{cer2018universal} to the SVM with RBF kernel for detecting hate speech in English HatEval. Their approach surpassed Neural Network models (such as FastText + BiGRU, BERT + CNN and GloVe + LSTM) because they stood at first rank.

By advancing and conducting various research in NLP, it is proven that Deep Neural Networks (DNNs) are capable of outperforming in compared to these traditional approaches; however, there are needed sufficient training instances for reaching optimum weights. \citeauthor{deeplearning_approaches} \citeyearpar{deeplearning_approaches} applied several Deep Learning approaches for detecting hate speech and concluded that exploiting ``LSTM + Random Embedding + Gradient Boosted  Decision Tree'' can outperform other methods. \citeauthor{gamback-sikdar-2017-using} \citeyearpar{gamback-sikdar-2017-using} applied several possible embedding approaches for feeding to a Convolutional Neural Network (CNN), which reached their best experimental result by exploiting Word2Vec. \citeauthor{park-fung-2017-one} \citeyearpar{park-fung-2017-one} proposed a model, named HybridCNN, for classifying hate speech, which takes both character and word features as input. \citeauthor{aroyehun-gelbukh-2018-aggression} \citeyearpar{aroyehun-gelbukh-2018-aggression} used FastText to represent inputs and use LSTM for classifying aggression in texts. They stood at the first rank in TRAC (Trolling, Aggression and Cyberbullying) shared task.

The emergence of the transformers and applying them for extracting features causes proposing state-of-the-art models in NLP tasks. Thanks to OffensEval shared tasks, that has been caused proposing many transformer-based approaches by participants. NULI \citep{liu-etal-2019-nuli}, UM-IU@LING \citep{zhu-etal-2019-um} and Embeddia \citep{pelicon-etal-2019-embeddia} are those examples of participants in OffensEval 2019 who fine-tuned BERT with variations in the parameters and preprocessing steps for subtask-A, which aimed to detect offensive language. UHH-LT \citep{wiedemann-etal-2020-uhh} fine-tuned different transformer models on the OLID training data, and then combined them into an ensemble. Galileo \citep{galileo} and Rouges \citep{dadu-pant-2020-team} are those participants who fine-tuned XLM-R (XLM-RoBERTa) to detect the offensive language in tweets across five languages (English, Turkish, Greek, Arabic, and Danish). XLM-RoBERTa is an unsupervised cross-lingual representation pretrained transformer model, which is highly scalable and can be fine-tuned for different downstream tasks in multiple languages \citep{xlmr}. GUIR \citep{sotudeh-etal-2020-guir} trained linear SVM classifiers (view-classifiers) using 6-gram features in addition to the BERT-based classifier and fed the concatenation of the probability output from SVM and sigmoid output from BERT as feature vector for a final linear SVM classifier (the meta-classifier). Kungfupanda \citep{dai2020kungfupanda} stood at the sixth rank in the English OffensEval 2020. They fed the text to the BERT model and passed its output to three similar levels of BiLSTM layers, followed by linear layers with different units. KUISAIL \citep{safaya-etal-2020-kuisail} showed that combining CNN
with BERT is better than using BERT on its own. This participant stood at top-five participants for Arabic, Greek and Turkish languages in OffensEval 2020 by using language-specific variations of BERT (separately for each language) and passed their output to a CNN, which is designed based on mixture of experts \citep{BALDACCHINO2016178} concept.

\subsection{Related Datasets}

The OLID 2019 is a hierarchical dataset for detecting the offensive language (Subtask-A), categorizing it (Subtask-B), and identifying its target (Subtask-C). Subtask-A is a binary classification problem that contains `NOT' and `OFF' labels. `NOT' refers to those posts, which do not contain any offense, whilst `OFF' determines posts including insults, threats, and posts containing profane language or swear words. The SOLID 2020 \citep{rosenthal2020large}, refers to Semi-supervised OLID, which were used several NLP techniques to prepare more volume dataset.

HatEval provided a hierarchical dataset, similar to OLID 2019. This dataset aims to identify hate speech against women and immigrants. The dataset is distributed among two languages, namely English and Spanish. The first level of English language version judges whether the content of the text is hatefully provided that `HS' (a binary value) becomes 1.

The TRAC-1 dataset \citep{KUMAR18.861}, has English and Hindi formats, in which the instances are categorized into 3 different labels, namely `NAG'/Not-Aggressive, `CAG'/Covertly Aggressive, `OAG'/Overtly Aggressive. \citeauthor{trac2-dataset} \citeyearpar{trac2-dataset} provides the second version of TRAC-1 and adds a second level for identifying misogyny. The second level is a binary classification, in which the gendered or misogynous contents are specified by `GEN', otherwise `NGEN' is used.

\section{Methodology}
\begin{figure}[]
  \centering
  \includegraphics[scale=0.65]{{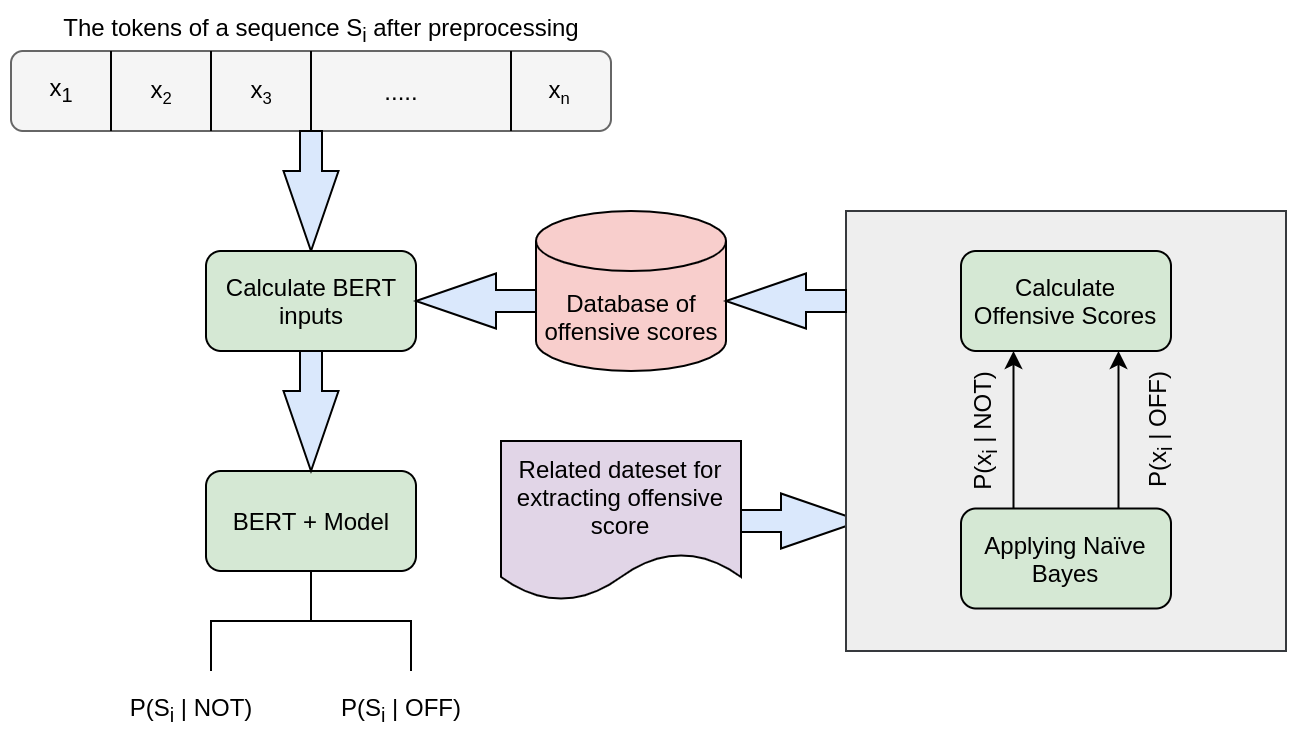}}
  \caption{Schema of the proposed method.}
  \label{fig:figure1}
\end{figure}

In this section, we present our methodology for detecting offensive contents in Persian and English languages. For this purpose, we first talk about the datasets including our newly created dataset, named POLID (for Persian OLID) and other exploited resources . Then, we describe our proposed models and algorithms. The overall view of our methodology is depicted in Figure \ref{fig:figure1}.

\subsection{Creating Persian OLID (POLID)}
To the best of our knowledge, there are not available and open datasets related to offensive language detection in Persian. Hence, we had to create our dataset. Due to collecting text data, we crawl tweets, Instagram comments, users’ reviews on different Iranian web applications such as Digikala \footnote{\url{https://www.digikala.com/}}, Snappfood \footnote{\url{https://snappfood.ir/}}, etc. We categorize the data in two steps, to make confident labels. At the first step, we create a basic list of common swearwords and label each text entity (text entity means each tweet, Instagram comment and users' review in Iranian web applications) as `OFF' if it contains at least one element of this list. Otherwise, we categorize it as `NOT'. The meaning of the labels is as follow: `NOT' determines those texts, which neither contain profanity, nor hating, nor threatening. On the other hand, `OFF' texts contain profanity, or hating, or threatening. In the second step, we correct the miss-labelled instances based on the following:

\begin{itemize}
    \item There are some Persian polysemous words, for which there are offensive and non-offensive senses; for instance, `` \FR{لخت} '' means naked by pronouncing `lokht' and inertia by pronouncing `lakht'. Hence, we may have labelled wrongly some instances as  offensive.

    \item The text data may contains offensive content with no obscenities. These types of text may contain targeted offenses, which may be veiled or direct. For instance, in the phrase of `` \FR{آخر سر میکشمت.} '' (Translation: I will kill you finally.), there are no swearwords, but it contains offensive language. Hence, we modified some instances, which were labeled as not offensive before.
\end{itemize}

The final dataset (POLID) contains 4,988 text entities, comprising 2,453 of not offensive and 2,535 of offensive contents.

\subsection{Employed Datasets}
There are two purposes for using the available datasets; 1) Extracting an underlying dictionary, which maps a token to its offensive score 2) Training and evaluating the models. In English, OLID, dataset of HatEval and TRAC-1 dataset were used for the first purpose and OLID was used just for the second purpose. In Persian, POLID dataset is applied for both purposes. Tables \ref{tab:1} and \ref{tab:2} illustrates the information about the distribution of used datasets for both purposes.

As discussed, each named dataset covers a particular aspect of offensive language,  hence it is logical to combine them and create more comprehensive dataset. To make our models comparable to others, we merely homogenized the named datasets for extracting the offensive scores and used OLID for training and evaluating the models. The homogenized dataset contains two labels, namely `NOT' and `OFF'. In  dataset of HatEval, 0s and 1s are considered as `NOT' and `OFF', respectively. In TRAC-1 dataset, NAGs are considered as `NOT', whilst CAGs and OAGs are considered as `OFF'.

\subsection{Preprocessing}

In English, we followed the preprocessing steps proposed by the Kungfupanda; however, there are some differences in Persian preprocessing steps. In continue, the modules and their differences will be discussed.

To deal with emojis, Kungfupanda replaces each emoji with its English textual descriptor. This is a logical approach for dealing with emojis because sometimes the combination of some emojis conveys an offensive concept and this information will not be missed by this module. But, in Persian,  there is no comprehensive library for replacing emojis with its Persian textual descriptor. Hence, we removed each emoji in the texts inevitably.

As both OLID and POLID entities have come from social media, there are many meaningful hashtags in them. Hence, it is more effective to extract the words from hashtags. For tackling this challenge, Kungfupanda exploited a third-party library. Since there is no Persian-based library, we first detect the hashtags, then by removing underlines we extract their words.

Due to reducing redundancy, Kungfupanda replaced all `@USER' with one `@USERS' at the first of the text, provided that the original text contains several `@USER'. On the contrary, we deleted users’ IDs and numbers to reduce redundancy because we focus on detecting offensive language not identifying the people who have been insulted.

Kungfupanda replaced `URL' tokens with `http' to handle Out-Of-Vocabulary (OOV) tokens. In Persian, we deleted the links because when we crawled Instagram comments, the text entities were comprising the hyperlinks to posts and they did not contain any useful information.

There are some further preprocessing steps that we applied on POLID. It includes normalizing elongated words and converting to proper Persian format. 

Users may repeat some letters in a specific word for representing their emotions, such as ` \FR{هوا خیلی خووووووووبه!!} ' (Translation: The weather is very gooooooood.). As the repetition does not affect our task, we normalized those tokens which have more than two consecutive repeated characters and lessened them to two repeated characters for reducing redundancy. The reason behind that is there are some words like ` \FR{گراییدن} ' (means to tend), which has two consecutive repeated characters and normalizing this word ruin its meaning. Finally, we normalize texts for character encodings by exploiting Hazm\footnote{\url{https://github.com/sobhe/hazm}}.

Eventually, similar to the last module of Kungfupanda's preprocessing, we truncated all the tweets to the max length of 64 to lower the GPU memory usage and slightly improve the performance.

\begin{table}
    \centering
    \caption{Distribution of datasets before homogenizing.}
    \begin{tabular}{c|cc|cc|ccc}
        \hline Dataset & \multicolumn{2}{|c|} { OLID } & \multicolumn{2}{c|} { HatEval } & \multicolumn{3}{c} { TRAC-1 } \\
        
        \hline Label & NOT & OFF & $\mathrm{HS}=0$ & $\mathrm{HS}=1$ & NAG & CAG & OAG \\
        
        \hline Counts & 8,840 & 4,440 & 5,217 & 3,783 & 5,052 & 4,240 & 2,078 \\
        
        \hline
    \end{tabular}
    
    \label{tab:1}
\end{table}

\begin{table}
    \centering
    \caption{The distribution of OLID 2019, which is used for training and evaluating the models.}
    \begin{tabular}{ccccccc}
        \hline \multicolumn{3}{c} { Training } & & \multicolumn{3}{c} { Testing } \\
        
        \cline { 1 - 3 } \cline { 5 - 7 } NOT & OFF & Total & & NOT & OFF & Total \\
        
        \cline { 1 - 3 } \cline { 5 - 7 } 8,840 & 4,400 & 13,240 & & 620 & 240 & 860 \\
        
        \hline
    \end{tabular}
    
    \label{tab:2}        
\end{table}

\subsection{Calculating Offensive Scores}
In this section, we present the approach of mapping tokens to a number between 0 and 1 as their offensive scores. The closer to 1 represents the more offensive a token is. As manual mapping is prone to error and it is possible to miss some offensive tokens, we assumed that the homogenized dataset (In English) and POLID (In Persian) are multinomially distributed and applied Multinomial Naïve Bayes for automatizing the procedure.

Forasmuch as the offensive scores should be calculated as what the BERT uses, we tokenized the text entities with the BERT tokenizer that Huggingface \citep{DBLP:journals/corr/abs-1910-03771} has implemented. Then, we applied Term Frequency Inverse Document Frequency (TF-IDF) and Multinomial Naïve Bayes to extract probabilities belonging to `NOT' and `OFF' classes for each token. These probabilities are calculated by equation \ref{eq:naive_bayes}, where $\hat{\theta}_{y i}$ is the probability $P(x_{i} | y)$
of token $i$ appearing in a sample belonging to class $y$ (in this case $y$ is `NOT' or `OFF'),  $N_{y i}$ is the number of times that token $i$ appears in a sample of class $y$ in the related dataset, $N_{y}$ is the total count of all features for class $y$ and $|V|$ is the size of the vocabulary.

\begin{equation}
    \label{eq:naive_bayes}
    \hat{\theta}_{y i} = \frac{N_{y i} + 1}{N_{y} + |V|}
\end{equation}

After calculating probabilities $P(x_i | NOT)$ and $P(x_i | OFF)$ for each token like $x_i$, we applied equations \ref{eq:equation1} and \ref{eq:equation2} to extract offensive score for each token. Table \ref{tab:4} illustrates some tokens that mapped to their offensive score. For English we used text entities from homogenized dataset, while for Persian whole entities of POLID were used. 

\begin{equation}
    \label{eq:equation1}
    S(x) = \frac{1}{1 + {e}^{-x}}
\end{equation}

\begin{equation}
    \label{eq:equation2}
    OffensiveScore(x_{i}) = S(log\frac{P(x_i | OFF)}{P(x_i | NOT)})
\end{equation}

\begin{table}
 \caption{Some instances of token's offensive score}
  \centering
  \begin{tabular}{l c l c}
    \toprule
    Token     & Offensive Score  \\
    \midrule
    idiots & 0.8843 \\
    sucks & 0.8352 \\
    filthy & 0.8316 \\
    pig & 0.8044 \\
    cows & 0.8043 \\
    stubborn & 0.6292 \\
    devil & 0.6225 \\
    disabilities & 0.4464 \\
    airport & 0.2489 \\ 
    \bottomrule
  \end{tabular}
  \label{tab:4}
\end{table}

These offensive scores will be saved in a database for future usage. Due to optimizing the performance of BERT or ParsBERT, the offensive scores will be applied in customizing the attention probabilities.

\subsection{Calculating BERT Inputs}
We used BERT and ParsBERT transformers, provided by Huggingface that get `Input IDs' and `Attention Mask' as required and optional parameters, respectively. `Input IDs' is a vector containing indices of input sequence tokens in the vocabulary. On the other hand, `Attention Mask' indicates which tokens should be attended to, and which should not. Classic `Attention Mask' contains 0s and 1s, which the value of padding tokens are 0 and the others are 1.

Equation \ref{eq:equation3} represents the approach of using the `Attention Mask' vector in the process of attention probabilities calculation. This equation is based on the implementation of BERT provided by Google Research\footnote{\url{https://github.com/google-research/bert/blob/eedf5716ce1268e56f0a50264a88cafad334ac61/modeling.py\#L705}}. Based on this equation, the complement of `Attention Mask' is multiplied by a very small number (like -10,000) and adds to pre-calculated attention probabilities. Then, new attention probabilities will be normalized by applying softmax. This causes to have less attention to padding tokens.

\begin{equation}
    \label{eq:equation3}
    AttentionProbabilities = Softmax(AttentionScores - 10000 * (1 - attention\_mask))
\end{equation}

Although the value of attention probabilities would be different for different tokens by exploiting the classic `Attention Mask', it is possible to be created such values that cause the models to not pay more attention to offensive phrases. Accordingly, we believe that the process of creating word embeddings (by BERT or ParsBERT) would be improved provided that attention probabilities were customized, which it can be done by creating a custom-tailored `attention mask' that focuses on offensive phrases more than others. For creating this new `Attention Mask', we recommend two possible equations such as equations \ref{eq:equation5}, \ref{eq:equation6} for calculating `Attention Mask' value for a single token $x_{i}$.

All proposed equations, keep the masking value of padding tokens as what it is was before. Their difference is in the `attention mask' value for the primary tokens. Equation \ref{eq:equation5} aims to add the offensive score of each token, provided that it finds the token in its database of the offensive score. It keeps the original `Attention Mask' value for those tokens that are not in the database. Actually, the model's attention to offensive phrases increases. On the other hand, equation \ref{eq:equation6} follows a different purpose. It aims to make the model just paying attention to very offensive tokens.

\begin{equation}
    \label{eq:equation5}
    AttentionMaskValue_{1}(x_{i}) =
    \begin{cases}
     0 & \text{if $x_{i}$ = padding token} \\
     1 + OffensiveScore(x_{i}) & \text{if $x_{i}$ in database} \\ 
     1 & \text{otherwise}
    \end{cases} 
\end{equation}

\begin{equation}
    \label{eq:equation6}
    AttentionMaskValue_{2}(x_{i}) =
    \begin{cases}
     0 & \text{if $x_{i}$ = padding token} \\
     1 & \text{if $OffensiveScore(x_{i})$ >= threshold} \\ 
     0 & \text{otherwise}
    \end{cases}   
\end{equation}

We applied both of these equations with various values of hyper-parameter (threshold in equation \ref{eq:equation6}) in both languages. Due to discovering the optimum value of the threshold, we applied the greedy search concept and set the value of the threshold between 0.5 and 0.8. If we set the value of threshold less than 0.5, we involve the tokens that $P(x_i | NOT) >= P(x_i | OFF)$ which means involving less offensive phrases that are not effective. Our best experimental result demonstrates that the best value of the threshold is 0.6.

Using the equation \ref{eq:equation6} is not a good choice in English models because we use various types of datasets to cover all aspects of offensive language when calculating offensive scores, which leads to having a small number of tokens with an offensive score greater than 0.6; consequently, paying attention to only a small number of tokens does not improve the performance of the models, but also confuses the models and reduces the performance.

On the contrary, using the equation \ref{eq:equation5} in Persian models does not change the performance of them. This is because the offensive text entities of the POLID contains profanity and aggressive behaviour more than covertly offensive language; consequently, the number of tokens with a score greater than 0.6 becomes greater (This number for English and Persian is respectively 3,352 and 5,588). Furthermore, in Persian, the tokens in this range (greater than 0.6) are more offensive, whilst this is not happened in English because there are some normal tokens with high offensive score; in result, using equation \ref{eq:equation5} makes the model not feel much difference between the offensive and normal tokens.

\subsection{Designing BERT-based Model}
Since the inputs of BERT (and ParsBERT) were prepared, we combined BERT with ANNs, CNNs and RNNs to analyze the effect of the customized attention probability in the performance of models. We used CNNs either because the result of using CNNs at NLP tasks were promising \citep{kim-2014-convolutional}\citep{article1234}. 

In English, we selected the re-implemented proposed models by top-performing participants of OffensEval 2019 and 2020, named of NULI, KUISAIL and Kungfupanda as the representatives BERT-based models of ANNs, CNNs and RNNs. These proposed architectures is introduced in Related Works section. We re-implement the architecture of Kungfupanda's model and use just their first level because our focus is in subtask-A of OLID.

In Persian, we define our models since there is not previous works. In these models, BERT is substituted with ParsBERT, whose output will be passed to some prevalent RNN and CNN models. The details of these models are as below:

\paragraph{ParsBERT + BiLSTM or BiGRU,} A Dropout layer after ParsBERT with rate of 0.5, followed by a Bidirectional LSTM or GRU layer with 32 units and a Dense layer for creating probabilities and classification.

\paragraph{ParsBERT + CNN\_BiLSTM,} A one-dimensional convolutional layer with 256 filters with size of 5, followed by a Max-Pooling layer with size of 2. The output is passed to a Bidirectional LSTM layer with 32 units. Batch Normalization layers is applied after Max-Pooling and BiLSTM layers for stabilizing the learning process and dramatically reducing the number of training epochs. Eventually two level of Dense layers with 16 and 2 units is used for classifying.

\paragraph{ParsBERT + CNN\_2D,} The idea is similar to the proposed model by KUISAIL. The difference is reducing the number of convolutional layers and adding a Batch Normalization layer after each of them. This is because the number of instances in the POLID dataset is small and training a model with high trainable parameters may cause overfitting.  The convolutional layers have 32 kernels of size 3x3, 64 kernels of size 4x4, 256 kernels of size 5x5. Finally, a dense layer with 2 units is added for classification. The new CNN\_2D possess nearly 1,000,000 learning parameters less than the KUISAIL's model.

\section{Experiments}
In this section, we aim to represent our best evaluation results that are obtained during the testing phase for both English and Persian languages. Macro-average F1 score is used for evaluating the performance of defined models, which were trained on training sets of OLID and POLID datasets. Furthermore, the effect of using offensive scores and the rate of improvement of the best models will be covered.

\subsection{English}
The experiments began with putting the re-implementation of the proposed model by NULI as our baseline model and fine-tune it with and without exploiting offensive scores. Then, further BERT-based models, such as the models proposed by Kungfupanda and KUISAIL were used due to having more confident conclusions and we repeated the training and testing processes. The models have been re-implemented in the TensorFlow \footnote{\url{https://www.tensorflow.org/}} framework and were trained on a single GeForce RTX 2080Ti.

Due to optimizing the weights of defined models during training phase, Adam Optimizer with learning rate of 1e-5 was used. The smaller values caused slow optimization and larger values caused the metric on the validation set not to be changed. The maximum number of epochs and batch size value were set 11 and 64, respectively. Table \ref{tab:5} represents the evaluation results on the testing set of OLID dataset, without and with using equation \ref{eq:equation5}.

\begin{center}
    \captionof{table}{Evaluation Results on re-implemented format of defined models.\label{tab:5}}
    
    \begin{tabular}{l | ccc | ccc  | c}
        \hline & \multicolumn{3}{|c|} { \textbf{Not Offensive}  } & \multicolumn{3}{|c|} { \textbf{Offensive}  } & \\
        
        \hline \textbf{Model}  & \multicolumn{1}{|c} {$\mathbf{P}$} & $\mathbf{R}$ & $\mathbf{F 1}$ & $\mathbf{P}$ & $\mathbf{R}$ & $\mathbf{F 1}$ & $\mathbf{F 1} \mathbf{~ M a c r o}$ \\
                
        \hline NULI & $0.8922$ & $0.9210$ & $0.9063$ & $0.7773$ & $0.7125$ & $0.7435$ & $0.8249$ \\
        
        NULI + equation \ref{eq:equation5} & $0.8915$ & $0.9274$ & $0.9091$ & $0.7907$ & $0.7083$ & $0.7473$ & $\mathbf{0.8282}$  \\
        
        \hline Kungfupanda & $0.8930$ & $0.9016$ & $0.8973$ & $0.7393$ & $0.7208$ & $0.7300$ & $0.8136$ \\
        
        Kungfupanda + equation \ref{eq:equation5} & $0.9042$ & $0.9129$ & $0.9085$ & $0.7692$ & $0.7500$ & $0.7595$ & $\mathbf{0.8340}$ \\
        
        \hline KUISAIL & $0.8717$ & $0.9532$ & $0.9106$ & $0.8407$ & $0.6375$ & $0.7251$ & $0.8179$ \\
        
        KUISAIL + equation \ref{eq:equation5} & $0.8877$ & $0.9306$ & $0.9087$ & $0.7952$ & $0.6958$ & $0.7422$ & $\mathbf{0.8254}$ \\
        
        \hline
    \end{tabular}
\end{center}

Our best model is the model proposed by Kungfupanda by applying equation \ref{eq:equation5}, which is achieved 0.8340 on the testing set of OLID. Based on reports, NULI achieved 0.829 of F1-Macro on the testing set of OLID, which our best model outperforms this model. Although Kungfupanda reports that their MTL\_Transformer\_LSTM model achieved the same score as our best model, the main difference between their model and ours is that they used the information of subtask-B and subtask-C during training, while we just used the information of subtask-A, which indicates that the proposed method has the positive impact on the performance.

\subsubsection{Error Analysis of the Best Model (Kungfupanda + equation \ref{eq:equation5})}

\begin{figure}
    \begin{subfigure}{.5\textwidth}
        \centering
        \includegraphics[width=.8\linewidth]{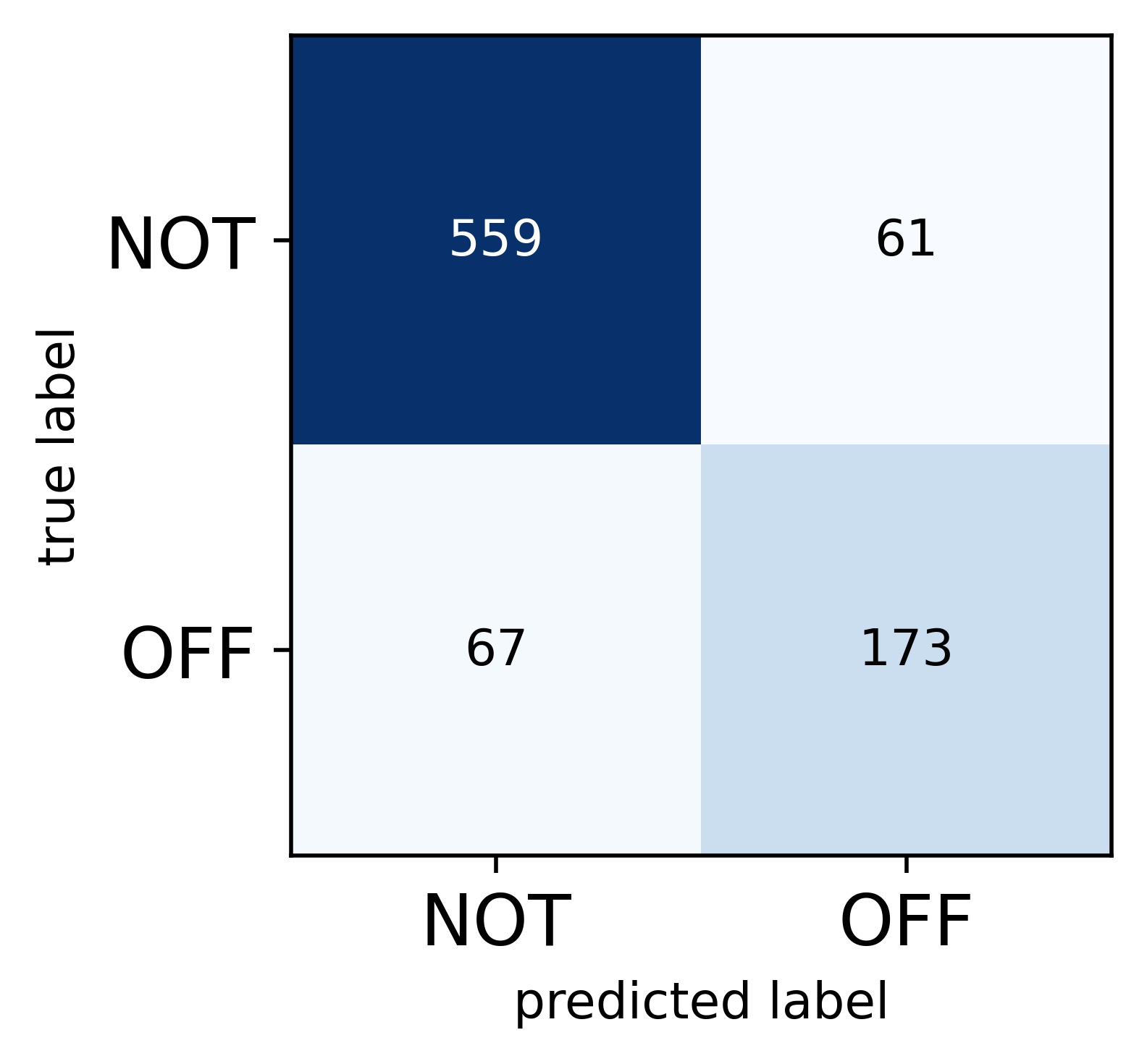}  
        \label{fig:sub-first}
    \end{subfigure}
    \begin{subfigure}{.5\textwidth}
        \centering
        \includegraphics[width=.8\linewidth]{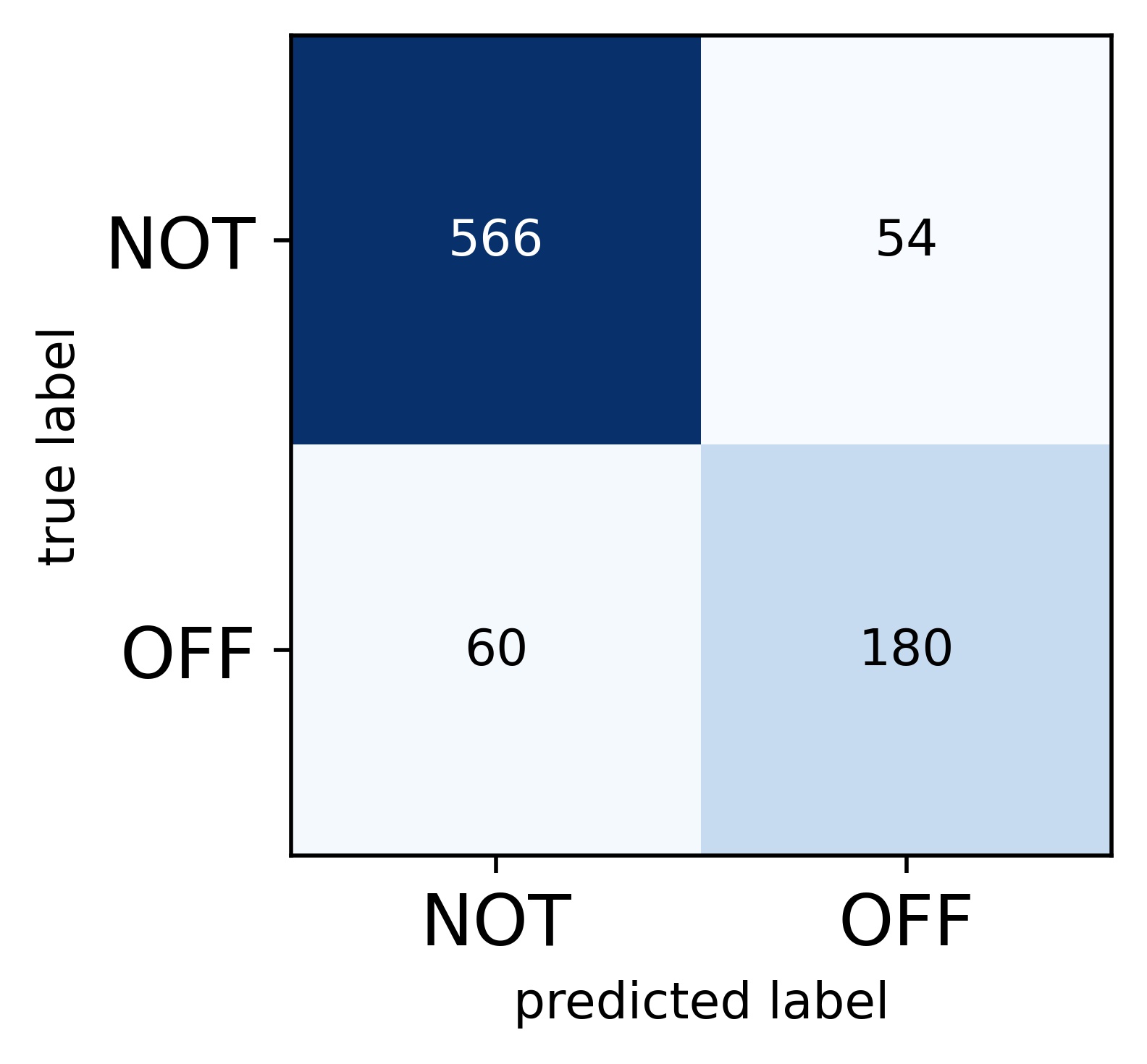}  
        \label{fig:sub-second}
    \end{subfigure}

    \caption{Confusion Matrix of the best experiments for English language, before and after using offensive scores. (a) the left is the confusion matrix of Kungfupanda without using equations. (b) the right is the confusion matrix of Kungfupanda by exploiting equation \ref{eq:equation5}.}
    \label{fig:fig}
\end{figure}

Figures \ref{fig:fig}a and \ref{fig:fig}b display the confusion matrices of the model proposed by Kungfupanda  (with or without exploiting equation \ref{eq:equation5}) against the testing set of OLID dataset. It can be seen that the enhanced model reduced both false positives and false negatives. 

The normal model classifies wrongly the text ``\#BeckyLynch is beautiful one of the few women in wrestling that absolutely need no work done at all. She's absolutely beautiful just the way she is. Doesn't need giant boobs or a fake booty. @USER is just simply amazing \#HIAC'' as an offensive language because of some phrases, such as `giant boobs' or `fake booty'. While, after exploiting equation \ref{eq:equation5} the amount of attention on different tokens will be changed that causes to predict this input as not offensive correctly. It turns out that the classifier, by re-distributing the attention mask, avoids over-focusing on some words that seem to be offensive at the first glance.

The enhanced model predicts the text ``\#Trump, everything trump does is another piece of evidence for obstruction of justice. He is a talking crime machine.'' as offensive. Adding the offensive scores of the tokens to the attention mask causes the model to predict this text as an offensive language with a probability of 0.92. While the normal model cannot understand the significance of `crime machine' and predicts it not offensive.

On the other hand, a few instances (18 text entities) cause the enhanced model to predict wrongly, while the normal model predicts these instances correctly. For instance, the text ``@USER This is one of my favourite quotes on the internet. He is so confident that he bends maths to his will. That's hardcore.'' is classified as offensive by enhanced model, while actually, it is not offensive. It is because of `hardcore' token, whose offensive score of 0.756 that causes the enhanced model to predict wrongly. As another example, ``\#AafiaSiddiqui is a convicted \#terrorist but she is widely hailed as the Daughter of \#Pakistan. URL'' is also classified as offensive by the enhanced model. which is wrong. This is because of phrases like `convicted terrorist', `daughter' and `\#Pakistan' that have high offensive scores and causes the model to predict this input as offensive wrongly.

\subsection{Persian}

Our Persian experiment began with preparing and labelling the dataset, named POLID. As the number of POLID instances is small and splitting it into train and test sets may cause missing some important features, we used Stratified K-fold, where $k = 5$ and calculate F1-Macro for each iteration and report the average of experiments in each fold.

We created a simple lexical-based classifier based on our manual swearwords list (an initial list, which was used for creating data) as our baseline model. This classifier categorizes each text entity as offensive provided that it contains at least one element of this list. The F1-Macro score of this classifier in 5 iterations of cross-validation is 0.6763, 0.6662, 0.6560, 0.6489 and 0.6880 (average: 0.6671). The high F1-Macro score for this baseline model indicates that the list covers most of the offensive instances.

Tables \ref{tab:6} and \ref{tab:7} report our best experimental results of models on different folds of POLID before and after applying equation \ref{eq:equation6}. As discussed, we applied different values as the threshold. The results showed us that 0.6 is the best value. Our best performance of the models archived by using the Adam Optimizer with learning rate of to 3e-4 for BiLSTM, 2e-5 for the model proposed by KUISAIL and 1e-3 for the others. Furthermore, The maximum number of epochs and batch size were set 30 and 64, respectively.

\begin{center}
    \captionof{table}{Evaluation Results on defined models before applying offensive scores.\label{tab:6}}
    \begin{tabular}{c|cc|cc|cc|cc|cc}
        \hline & \multicolumn{2}{|c|} { BiLSTM } & \multicolumn{2}{c|} { BiGRU } & \multicolumn{2}{c|} { CNN2D } & \multicolumn{2}{c|} { CNN-BiLSTM } & \multicolumn{2}{c} { KUISAIL } \\
        \hline & $\mathbf{Loss}$ & $\mathbf{F1}$ & $\mathbf{Loss}$ & $\mathbf{F1}$ & $\mathbf{Loss}$ & $\mathbf{F1}$ & $\mathbf{Loss}$ & $\mathbf{F1}$ & $\mathbf{Loss}$ & $\mathbf{F1}$ \\
        
        \hline $\mathbf{Fold 1}$ & $0.5436$ & $0.7515$ & $0.5456$ & $0.7474$ & $0.6490$ & $0.7605$ & $2.4684$ & $0.7485$ & $0.5099$ & $0.7662$ \\
        
        \hline $\mathbf{Fold 2}$ & $0.5524$ & $0.7493$ & $0.5232$ & $0.7665$ & $0.6335$ & $0.7709$ & $2.0774$ & $0.7652$ & $0.4810$ & $0.7685$ \\
        
        \hline $\mathbf{Fold 3}$ & $0.5285$ & $0.7444$ & $0.5482$ & $0.7509$ & $0.6646$ & $0.7605$ & $2.2875$ & $0.7484$ & $0.4915$ & $0.7675$ \\
        
        \hline $\mathbf{Fold 4}$ & $0.5015$ & $0.7670$ & $0.4952$ & $0.7773$ & $0.6130$ & $0.7733$ & $2.4643$ & $0.7560$ & $0.4775$ & $0.7752$ \\
        
        \hline $\mathbf{Fold 5}$ & $0.5468$ & $0.7408$ & $0.5301$ & $0.7472$ & $0.6404$ & $0.7561$ & $2.5533$ & $0.7382$ & $0.4990$ & $0.7440$ \\
        
        \hline $\mathbf{Average}$ & $0.5346$ & $0.7506$ & $0.5285$ & $0.7579$ & $0.6401$ & $0.7643$ & $2.3702$ & $0.7513$ & $0.4918$ & $0.7643$
    \end{tabular}
\end{center}

\begin{center}
    \captionof{table}{Evaluation Results on defined models after applying offensive scores\label{tab:7}}
    \begin{tabular}{c|cc|cc|cc|cc|cc}
        \hline & \multicolumn{2}{|c|} { BiLSTM } & \multicolumn{2}{c|} { BiGRU } & \multicolumn{2}{c|} { CNN2D } & \multicolumn{2}{c|} { CNN-BiLSTM } & \multicolumn{2}{c} { KUISAIL } \\
        \hline & $\mathbf{Loss}$ & $\mathbf{F1}$ & $\mathbf{Loss}$ & $\mathbf{F1}$ & $\mathbf{Loss}$ & $\mathbf{F1}$ & $\mathbf{Loss}$ & $\mathbf{F1}$ & $\mathbf{Loss}$ & $\mathbf{F1}$ \\
        
        \hline $\mathbf{Fold 1}$ & $0.4359$ & $0.8186$ & $0.3649$ & $0.8547$ & $0.4774$ & $0.8567$ & $2.1694$ & $0.8467$ & $0.3794$ & $0.8405$ \\
        
        \hline $\mathbf{Fold 2}$ & $0.4223$ & $0.8276$ & $0.3708$ & $0.8475$ & $0.4629$ & $0.8557$ & $1.7839$ & $0.8477$ & $0.3590$ & $0.8437$ \\
        
        \hline $\mathbf{Fold 3}$ & $0.3963$ & $0.8295$ & $0.3681$ & $0.8435$ & $0.4521$ & $0.8676$ & $2.0818$ & $0.8427$ & $0.3513$ & $0.8517$ \\
        
        \hline $\mathbf{Fold 4}$ & $0.3802$ & $0.8533$ & $0.3238$ & $0.8785$ & $0.4206$ & $0.8836$ & $2.3822$ & $0.8551$ & $0.3354$ & $0.8696$ \\
        
        \hline $\mathbf{Fold 5}$ & $0.4602$ & $0.8204$ & $0.3960$ & $0.8385$ & $0.4954$ & $0.8515$ & $2.3775$ & $0.8265$ & $0.3933$ & $0.8325$ \\
        
        \hline $\mathbf{Average}$ & $0.4190$ & $\mathbf{0.8299}$ & $0.3647$ & $\mathbf{0.8525}$ & $0.4617$ & $\mathbf{0.8630}$ & $2.1590$ & $\mathbf{0.8437}$ & $0.3636$ & $\mathbf{0.8476}$ \\
        
    \end{tabular}

\end{center}

\subsubsection{Error Analysis of the Best Model (CNN2D + equation \ref{eq:equation6})}

There are many instances that normal CNN\_2D fails at predicting the instances of the validation set in folds, while the enhanced one predicts correctly; for instance, the input of ``\FR{اگر میخواید که یک مارمولک و موزمار باشید، سلنا باشید} (Translation: If you want to be a person like lizards and muzmaars (a Persian idiom which means cunning and insidious), be like Selena.)'' is classified as a not offensive language by normal CNN\_2D. By taking a deeper look at this input, it can be understood that the `\FR{موزمار}' token is the principal cause of becoming offensive. This token has an offensive score of 0.7, which cause the enhanced model to pay attention to this token and classifies this input as offensive. As another example, the input `` \FR{جنسی کاملا معمولی و اندازش کوچکتر از عکس} (Translation: a normal production and smaller size than the photo)'' is not offensive actually but the normal CNN\_2D predicts it as an offensive language. The enhanced model figures out that this input does not have any offensive language; however, it criticizes a little bit.

On the other hand, there are some instances that the enhanced model fails at predicting correctly; for instance, the enhanced model fails at predicting ``\FR{چرت و پرت نگو ضعیفه. دلیلشو که نگفت، فقط نتیجه رو گفت. تو هم داری مثلا دلیلشو میگی، نتیجه بازم فرق نکرد.} (Translation: don't say bullshit weak lady, he didn't say the reason, he just said the results. You are talking about its reason but the results have not been changed.)'' correctly because of two reasons: 1) At this time, the model did not understand that ` \FR{چرت و پرت} ' has an inappropriate meaning. 2) Multinomial Naive Bayes could not extract appropriate probabilities for `\FR{ضعیفه}' which is a hateful word against women because its offensive score is near 0.53.

\section{Conclusion and Future works}
As presented, paying more attention to more offensive phrases than others can improve the performance of the models in identifying insulting language because these phrases have more impact on the final target. To do so, we created a new `Attention Mask` input by using a term named offensive score to help the BERT-based models more effectively detect offensive content. Although applying this term is promising, mapping a proper value to each token is challenging. Hence, we used a statistical approach like Multinomial Naive Bayes to find offensive scores automatically.

In the future, increasing the instance of POLID will be considered. Therefore, using a semi-supervised or unsupervised technique to find offensive scores and classify textual entities can be done because labelling all items will become time-consuming. Also, one thing that can enhance the performance is improving preprocessing modules, such as converting emojis to their Persian textual descriptors.

In addition, applying such techniques which consider language structures may improve the efficiency of this methodology. Moreover, exploiting other approaches for finding probabilities (for offensive scores extraction), such as Deep Learning methods, may improve the performance of the proposed methodology. On the other hand, exploiting this approach on multi-label classification problems may be promising.

\bibliography{references}

\end{document}